# Human-level molecular optimization driven by mol-gene evolution

Jiebin Fang,[1,2,⊥] Churu Mao,[2,⊥] Yuchen Zhu,[3,⊥] Xiaoming Chen,[2] Chang-Yu Hsieh[3]* & Zhongjun Ma[1,2]*

[1]Hainan Institute of Zhejiang University, Sanya, 572025, China
[2]Institute of Marine Biology and Pharmacology, Ocean College, Zhejiang University, Zhoushan, 316021, China
[3]College of Pharmaceutical Sciences and Cancer Center, Zhejiang University, Hangzhou 310058, Zhejiang, China
*Email: mazj@zju.edu.cn (Z.M.); kimhsieh@zju.edu.cn (C.H.)
⊥These authors contributed equally

**ABSTRACT:** *De novo* molecule generation allows the search for more drug-like hits across a vast chemical space. However, lead optimization is still required, and the process of optimizing molecular structures faces the challenge of balancing structural novelty with pharmacological properties. This study introduces the Deep Genetic Molecular Modification Algorithm (DGMM), which brings structure modification to the level of medicinal chemists. A discrete variational autoencoder (D-VAE) is used in DGMM to encode molecules as quantization code, mol-gene, which incorporates deep learning into genetic algorithms for flexible structural optimization. The mol-gene allows for the discovery of pharmacologically similar but structurally distinct compounds, and reveals the trade-offs of structural optimization in drug discovery.

The process of drug structure discovery can be divided into two main stages: hit discovery and lead optimization[1]. Hit compounds can be obtained through experimental testing of chemical entities[2], such as natural products[3], or by screening a large number of chemical structures using computational methods[4]. However, compounds obtained from a single screening are usually insufficient to meet the multiple requirements of drug properties. Therefore, the molecules need to undergo multiple steps of structural modification until a suitable lead molecule is obtained. This modification process involves complex structural changes and requires medicinal chemists to invest significant effort in designing and selecting the molecular structure based on their experience.

Computational methods can expand chemical space and aid in the discovery of more drug-like hit compounds, reducing the cost of structure optimization[5]. For instance, V-SYNTHES generates libraries of billions of compounds by combining pre-enumerated fragments[6]. Hit compounds are identified through screening using rule-based methods, such as structural alerts, and cheminformatics scoring methods. Genetic algorithms[7] or reinforcement learning[8] can optimize the combination strategy of fragments, reducing computational costs and considering more molecular properties. However, these algorithms are constrained to search only a limited range of structures by pre-enumerated fragments. The *De Novo* structure generation method, based on deep learning, is capable of generating entirely new chemical structures. By sampling the restricted potential space using VAE, DDR1 kinase inhibitors were successfully found[9]. Reinforcement learning algorithms, such as REINVENT[10] and MolDQN[11], have the ability to generate drug-like molecules or improve drug-like properties. However, this can come at the cost of reduced molecular diversity. Recent diffusion models enable the design of novel molecules[12] and also the generation of ligands in protein pockets[13], but the resulting molecular structures lack pharmaceutical properties.

By examining the actual drug discovery process, we have identified the following traits: The designed structure is typically based on an existing structure, such as a natural product or active fragment[3]. Multiple rounds of structure optimization are performed to obtain the best possible structures, and the optimization strategy is adjusted based on the results of each round[14]. The process of molecular optimization is not limited to the original structure, and the resulting compound may have a similar binding pattern to the reference molecule, but be structurally different[15,16]. Previous genetic algorithms utilized deep learning models for end-to-end molecular modifications[17], ignoring the importance of motifs to drug molecules. Modof[18] can modify molecular fragments, but does not treat the molecular structure as a whole and cannot perform structural modifications like bioequivalent substitutions. Therefore, we propose a Deep Genetic Molecule Modification Model (DGMM), which encodes molecules through deep learning as a basis for genetic algorithm (GA) modification. Variational auto-encoder (VAE)[19] can encode SMILES[20, 21] or molecular graphs[22] to obtain the latent chemical space and use it for downstream tasks. However, the continuous latent space obtained cannot be fully utilized due to the difficulties in interpretability. Discrete latent codes can be used for genetic operators but are inadequate for characterizing molecules as they ignore the details of the latent code space[23]. Hence, we use a VAE model to learn a distribution that contains molecular structure information. Subsequently, we distill the pre-trained VAE model into discrete variational auto-encoder (D-VAE) model to extract the quantized feature vectors of molecules, which we term 'mol-gene'. The mol-gene was evaluated and shown to effectively control molecular generation. In mol-gene space, we observed a distinction between molecular activity and drug-like properties. In the application of inhibitor discovery, DGMM's scaffold hopping and motif growth processes were found to be comparable to those of a medicinal chemist. DGMM distinguishes the chirality of the carbon interacting with the key site, further illustrating the algorithm's search capability. We applied the algorithm in a real drug discovery process and confirmed the validity of the resulting molecules.

## Results

### Design of DGMM

DGMM comprises of D-VAE and GA. D-VAE encodes molecules as discrete code, mol-gene (Fig. 1a), while GA performs structural optimization based on mol-gene (Fig. 1b). When training molecular coding models, codebook collapse happens when molecules are discretely encoded directly. This is because small molecule drugs usually contain only a limited number of common atoms, such as C, H, O, N, S, P, F, Cl, Br, and I. The use of graph convolution[24] or message-passing networks[25] could potentially improve this issue. However, considering the large number of encoding and decoding operations required for the subsequent GA, we use SMILES to represent the molecules. In order to enhance the model's focus on the ring and branching information of the molecule, we pre-code the SMILES of the molecules using SELF-referencIng Embedded Strings (SELFIES)[26]. Subsequently, the D-VAE model was developed to encode SELFIES, in accordance with previous studies which have demonstrated the effectiveness of the VAE model in extracting molecular features[20, 22]. The D-VAE model was obtained through three steps. Firstly, we trained the Teacher VAE model (T-VAE) to encode SELFIES as continuous feature vectors and reduce them to SELFIES. Inspired by the hierarchical control of the Style-GAN[27] for generating results, we implemented hierarchical feature generation in the T-VAE decoder to enhance result detail. Secondly, the T-VAE encoder is distilled into the D-VAE encoder using knowledge distillation[28]. The workflow of D-VAE involves quantizing continuous feature vectors and then reducing them back to continuous variables based on a codebook[23]. However, the decoder trained at this point is not a true decoder. The D-VAE encoder encodes SELFIES as mol-gene, which are then transformed into continuous feature vectors by a multilayer perceptron (MLP) decoder. These vectors are then fed into the T-VAE decoder to obtain a reconstructed distribution. In the third step, the D-VAE encoder is frozen while the real D-VAE decoder is obtained by fine-tuning the T-VAE decoder according to the reconstruction loss.

Genetic algorithms start with a single molecule, known as an ancestor molecule, as the initial population. During an evolution epoch, the molecular population are encoded into mol-gene, which are then mutated and crossed over to produce offspring mol-genes. Then, instead of decoding the mol-genes directly, we decode the mol-gene on multiple occasions until the offspring molecules no longer fulfill any of the conditions that resulted in a selection score (S-score) of 0. Alternatively, the mol-gene is discarded after the maximum number of decodes has been reached. The molecules that are successfully expressed are screened based on a S-score, which was calculated from quantitative estimation of drug-likeness (QED)[29], synthesis accessibility (SA)[30], ancestral similarity and population diversity. Molecular competition is based on protein-ligand affinity energies (kcal/mol) calculated using force-field-based molecular docking[31, 32]. An evolutionary process consists of several epochs, and the top-ranking molecule in each epoch is called an "elite molecule" and it will be recorded. The maximum common structure (MCS) of the elite molecules was used to calculate the similarity, instead of the ancestor molecules, to achieve scaffold degradation. Additionally, the active groups of the elite molecules are collected, which will be the parameters for the S-score calculation (Fig. 1c).

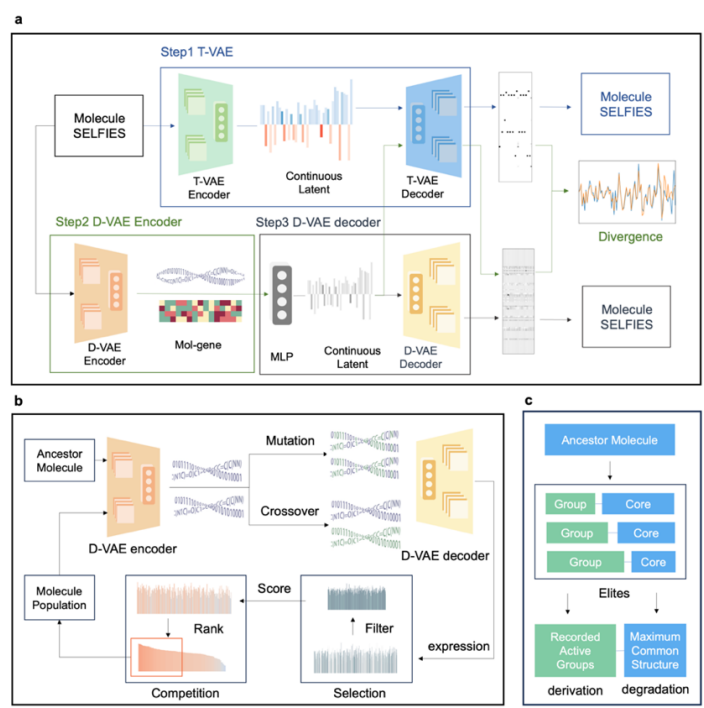

**Fig. 1 | DGMM overview. a.** D-VAE model. Step 2 is trained on the divergence of the T-VAE and D-VAE output distributions, while the other two steps rely mainly on the SELFIES reconstruction loss. Once each training step is completed, it will be frozen for the subsequent training. **b.** GA workflow. The genetic algorithm begins with the ancestor molecule, generates offspring using crossover, mutation and expression, and then selects the population through competitive screening. This process is repeated until a predetermined number of iterations is achieved. **c.** Derivation-degradation mechanism. At the beginning, the algorithm considers the entire ancestor molecule as the core, and then degrades the core according to the maximum common structure of the core and the elite molecules. The groups of the molecule are split according to the core. The groups in offspring elite molecules are recorded and used to further degrade the core.

## Application of mol-gene

The chemical spaces defined by molecular fingerprints and mol-gene are compared, and the distribution of molecules in each space is observed. Molecules are more centrally distributed in the fingerprint chemical space (Fig. 2a) than in the mol-gene chemical space (Fig. 2b). Since the active molecules differ only subtly between substructures, it is difficult to distinguish between compounds of different structural types in the fingerprint chemical space, whereas the mol-gene space is able to do so. Moreover, the outlier molecules in the mol-gene clustering are structurally different from the other molecules. The mol-gene mutation results indicate that the mol-gene are correlated with molecular structural features. Mutations in mol-gene essentially do not the conjugated scaffold (benzene), but rather the substituents (Fig. 2d). Subsequently, the crossover is employed to introduce external information. As illustrated in Fig. 2e, the Cross-S-0 mutation generates molecules with electron withdrawing group (EWG) substituents on the benzene ring. Conversely, the Cross-S-1 mutation produces molecules with electron donating group (EDG) substituents. When the mol-gene of Cross-S-0 is crossed with that of Cross-S-1, the resulting molecule maintains the Cross-S-0 scaffold, but with EDG substituents, as in Cross-S-1.

The distributions of QED and SA in the mol-gene chemical space are clearly clustered upon analyzing the clustering of the molecular properties of the mol-gene. When analyzing the activity of molecules and their drug-like properties, we observed that these two properties are distributed in opposite directions in the mol-gene space (Fig. 2c). This explains why the structural drug-like properties deteriorate as the molecular activity increases.

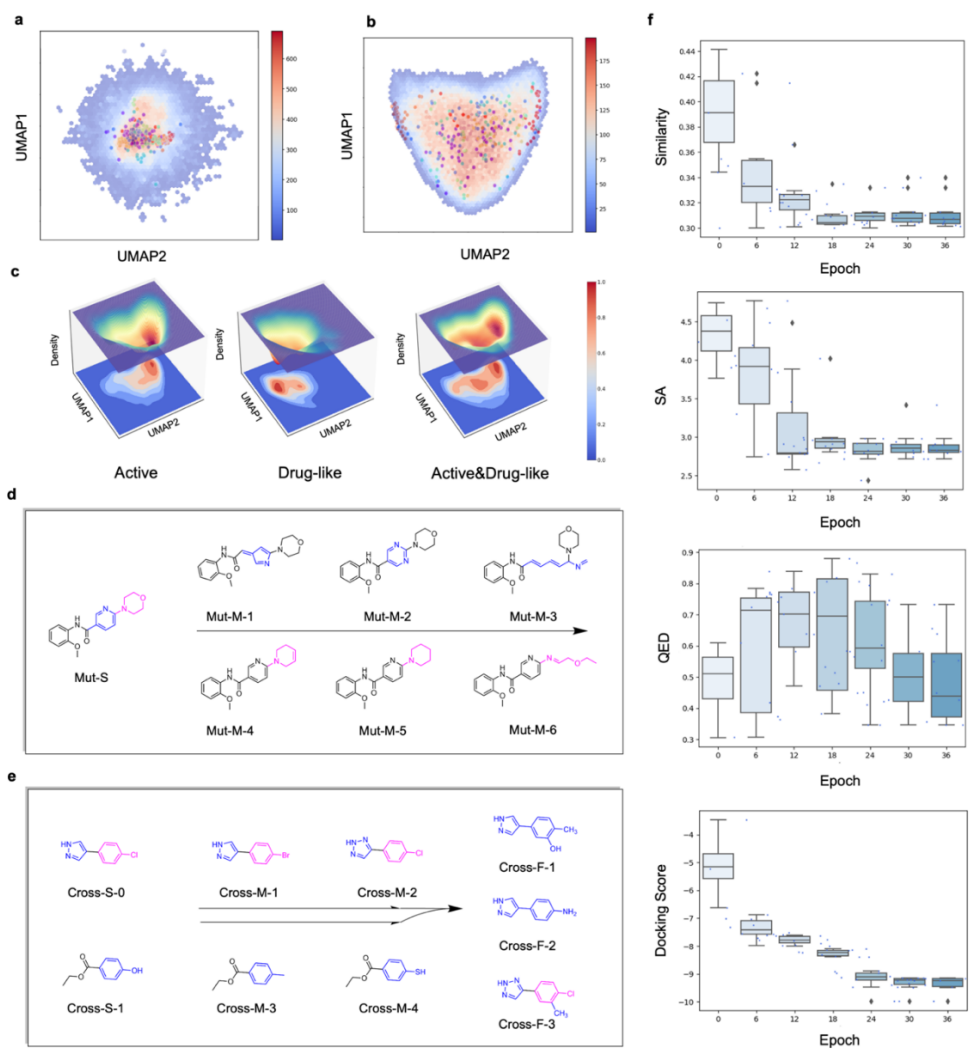

**Fig. 2 | Application of mol-gene.** UMAP distribution of ZINC database[33] compounds using **a.** molecular fingerprinting[34] and **b.** mol-gene. Scatters represent compounds with ROCK2 $IC_{50}$<1 μmol/L in BindingDB[35], and the color of the scatter indicates the structure type, which is classified by a medicinal chemist. **c.** The distribution of different molecules in the mol-gene. Active includes molecules with $IC_{50}$ values less than 0.1 μmol/L (regardless of target) in BindingDB, Drug-like includes molecules with QED>0.6 and SA<2.6 in the ZINC database, and Active&Drug-like includes molecules that meet both $IC_{50}$ < 1 μmol/L and Drug-like conditions. **d.** Mutations using mol-gene. Mut-S is the input structure and structures with the prefix Mut-M is the mutation output. **e.** Crossovers using mol-gene. Structures with the prefix Cross-S are input molecules, structures with the prefix Cross-M are the result of mutations of each input structure, and structures with the prefix Cross-F are the output from crossover. Test structures are obtained by random sampling in the ZINC database[33]. **e.** Evolution of molecular properties. A higher similarity indicates a closer structural resemblance to the starting molecule. A lower SA score indicates a simpler synthesis of the molecule, while a higher QED score indicates a molecule that is more similar to the drug in terms of its physicochemical properties[29, 30]. A lower docking score indicates better binding ability of the molecule to the ligand.

## Evaluation of DGMM

We use SELFIES as a precoding of the molecule, and all subsequently generated SMILES are syntactically and semantically valid, and the algorithm will only obtain molecules with desirable physicochemical properties by screening of molecular properties. Therefore, instead of evaluating the number and properties of molecules generated by the algorithm, we focus on the change in molecular properties between evolutionary epochs. During evolution, a gradual decrease in the docking score can be observed, eventually reaching a level close to -10 kcal/mol. DGMM-generated molecules have SA values that gradually decrease with evolution, eventually stabilizing essentially below 3, and there is no significant trend in the change of QED scores (Fig. 2f). During the process of molecule optimization, molecules with excellent pharmacological properties but low docking scores are eliminated from the competition, while the retention of molecules with poor pharmacological properties but high docking scores depends on the S-score.

Then, We compare the structure modification process of DGMM with that of medicinal chemistry experts. We examine the design of checkpoint kinase 1 (CHK1) inhibitor structures, which typically follow the classical structure-based drug design approach[14, 36]. It should be noted that the molecules derived from the algorithm are rich in diversity, and we will focus more on results that are similar to reported structures. As with the actual medicinal chemistry optimization[14], we applied the algorithm for derivatization using CHK1-S-0 as the starting input molecule. The algorithm produced CHK-F-1, a structure that closely resembles CHK1-S-1, a design by medicinal chemistry experts (Fig. 3a). Specifically, the modification process added additional rings to enlarge the pyrazolopyridine group and introduced nitrogen atom substitutions in the aromatic ring. This increases the affinity of the molecule for polar residues in the CHK1 protein binding pocket (which also improves the selectivity for CHK1 due to the hydrophobicity of the binding pocket of the CHK2 of the same family)[14].

When CHK1-S-0 is input, the algorithm produces multiple types of molecular structures. Thus, we reset the input to CHK1-S-1 to restrict the structural characterization of the molecular population. At this point, the CHK1-S-2 and CHK1-S-3 series molecules become the primary derivation directions. This indicates that when the structure type is determined, the algorithm can selectively disconnect the intermediate ring like a medicinal chemistry expert to improve structural flexibility.

We analyzed how the modifications are made to explore whether the algorithm can provide insights into structure-activity relationships. Molecule substructures that underwent significant changes during the evolution process are identified and weighted based on their impact on activity (Fig. 3b). The results show that the algorithm can guide molecular activity optimization. The plausibility of the structural evolution was verified through the binding mode obtained from molecular dynamics[37]. The algorithm preserves the two key hydrogen bonding interactions formed by the nitrogen atom with the residues Glu17 and Cys87, and controls the molecular size to match the space of about 9 Å within the binding pocket (Fig. 3c).

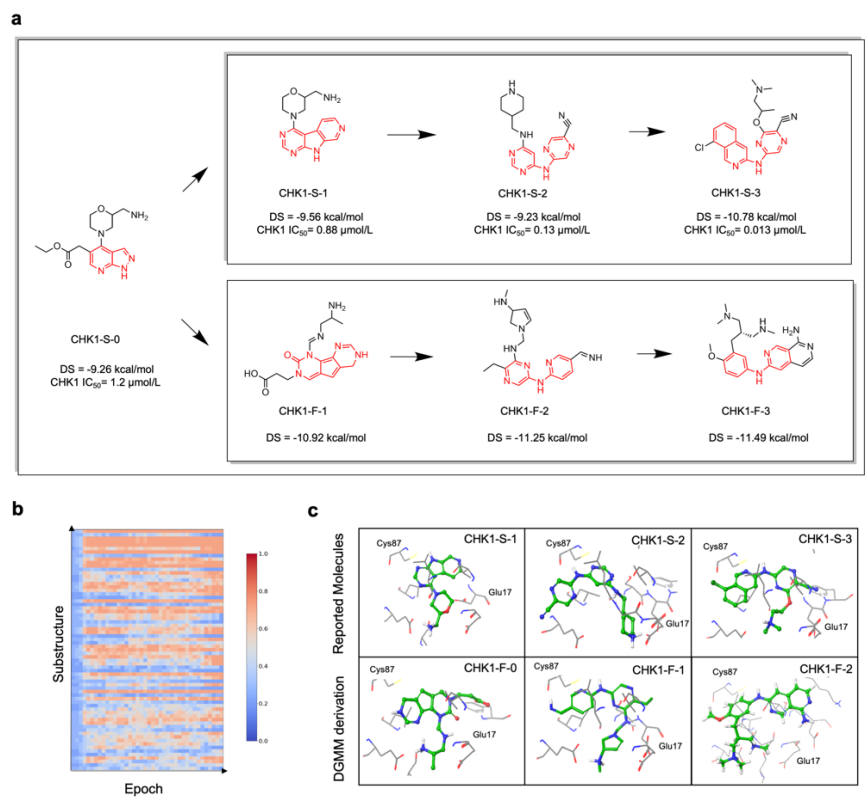

**Fig. 3 | CHK1 inhibitor discovery. a.** CHK1 inhibitor discovery. Molecules with the prefix CHK1-S are reported modifications and molecules with the prefix CHK1-F are the outputs of the model. **b.** Structural evolution analysis. Substructures represent a number of pre-listed structural fragments. Substructures that change more than 40% are filtered and displayed. The color represents the frequency of occurrence of the substructure, and the color change can be used to determine whether the substructure is retained (turns red) or eliminated (turns blue). **c.** CHK1 binding mode. The binding mode was obtained by induced fit docking (IFD)[38] of the compounds and the CHK1 protein (PDB: 5DLS).

## Discussion

In our research, we developed the DGMM algorithm to achieve molecular modifications by balancing the structural changes brought about by the mol-gene through genetic algorithms. The mol-gene gives the modification process a broad search range that converges to the appropriate chemical space with the help of genetic algorithms. The GA enables the derivation of new molecules that retain the characteristics of the original molecule while not being constrained by its original structure. This is a crucial aspect of lead compound optimization and helps in the discovery of potential drug molecules.

The challenge of achieving a balance between structural novelty and molecular properties is a general problem in the field of molecular modification, affecting medicinal chemistry practice and ultimately leading to limitations in computational methods. For instance, HDAC8 inhibitors are often developed using the hydroxamic acid moiety as the zinc binding group, due to its strong binding affinity[16]. However, the dominant hydroxamic acid limits the chemical space for derivatives[39]. These potential molecules, rich in structural diversity, will be overlooked due to preconceived pharmacochemical experience. In addition, limited experimental data can also constrain data-driven methods. Deep learning can be trapped by hydroxamic acid, making it difficult to access other molecules. We only apply deep learning to chemical structure derivation, leaving the search for suitable chemical spaces to molecular structure-based computational methods.

For molecular characterization, we have attempted to implement a variety of molecular descriptors, molecular fingerprints, and VAE as molecular encoding approaches, with VAE proving to be the most effective. This could be due to the VAE model combines similar structure patterns to form molecular distributions when generalizing over a large number of molecules, which aligns more closely with human medicinal chemistry experts. With mol-gene, it can be observed that the activity and physicochemical properties of the molecules are actually different structural trends. This can be explained by the fact that biologically active molecules can influence the normal cells, and therefore such molecules are resisted by organisms. As the DGMM molecular optimization progresses, the docking score increases to the active level and the QED remains at a certain level. During drug development, molecular physicochemical properties are not always ideal, as they can be improved by novel drug delivery modalities[40, 41]. Additionally, it is not always the case that compounds that bind effectively to proteins are also cellularly active[42]. Therefore, drug discovery is not exactly a multi-objective optimization problem, and multi-step optimization is a more promising approach than single-step generation for identifying molecules that meet the actual drug requirements.

Nevertheless, the DGMM algorithm still has limitations. Firstly, despite its capacity to support the vast majority of molecules, DGMM with SELFIES as the molecule input is unable to process molecules with sequence lengths that are too long or molecules that present tokens not pre-enumerated. Secondly, the direction of DGMM derivation is determined by the scoring method. While there is a strong correlation between molecular docking and real activity data, it is important to note that the two are not equivalent. The effectiveness of molecular docking is influenced by the process of protein structure in molecular docking. Besides, DGMM relies solely on molecular docking algorithms to evaluate the binding capability between molecules and target receptors, which indicates that the method is unsuitable for the development of target-binding molecules for which activity cannot be predicted using molecular docking. Finally, although we considered a variety of scoring methods for molecule selection, as previously stated, drug molecule modification is not exactly a multi-objective optimization problem, requiring trade-offs in properties. The selection of molecules for the wet experiment was made in collaboration with medicinal chemistry experts. Consequently, the optimal methodology for weighting the scores and selecting the molecules among the various types of structures optimized by the algorithm is yet to be determined through further research.

## Methods

### Molecule process and Dataset building

From the ZINC compound library of 9,221,059 compounds[33], molecules with SMILES lengths less than 80 were screened, and molecules and mixtures with charges were excluded, resulting in 9,199,890 molecules. All molecule SMILES are first converted into canonical SMILES and then into a randomized SMILES format. An alphabet of 120 tokens is constructed from all molecules in the library by encoding SMILES in SELFIES format, and a condensed dataset covering the entire alphabet of 188,200 molecules was obtained by gradual cumulative random sampling. In addition, the same data processing of 1,526,981 molecules from the BindingDB compound library yielded 1,396,472 active molecule compounds for active compound analysis[35]. The 167-bit MACCS molecular fingerprints computed by RDKIT using RDKFingerprint are used for the molecular substructure analysis. Similarity scores are obtained using the Tanimoto similarity calculation based on RDKFingerprint. Randomized SMILES are obtained by randomizing the order of atoms in the Mol objects and then converting them to SMILES[34].

### Architecture and Training of T-VAE

The loss function is given in Eq. 1. The first term of the loss function is the Kullback-Leibler (KL) Divergence loss of the VAE[19], $\sigma$ and $\mu$ are the variance and mean of the prior distribution, $L$ is the latent size, and $\gamma$ is the hyperparameter that adjusts the KL loss, and the variation of $\gamma$ with training period $n$ is given in Eq. 2. The last two terms of the loss function are the reconstruction loss based on Cross Entropy ($CE$) and the molecular fingerprint loss based on Binary Cross Entropy ($BCE$), where $x$ is the molecular input, $\hat{x}$ is the predicted input, $x_{fp}$ is the true molecular fingerprint, and $\hat{x}_{fp}$ is the predicted molecular fingerprint.

$$Loss_{T-VAE} = -\gamma * \frac{1}{2}\sum_{j=1}^{L}\left(1 + \log(\sigma_j^2) - \mu_j^2 - \sigma_j^2\right)$$
$$+ CE(x, \hat{x}) + BCE(x_{fp}, \hat{x}_{fp}) \tag{1}$$

$$\gamma^n = \begin{cases} \gamma^{n-1} + 10^{-5} & \text{if } n > 1 \text{ and } \gamma^{n-1} < 10^{-4} \\ 10^{-5} & \text{else} \end{cases} \tag{2}$$

### Architecture and Training of D-VAE

Two different encoders were tried for the D-VAE model. The loss function is given in Eq. 3. The loss function is computed from the cross entropy at temperature $T$, $\alpha$ is the hyperparameter for knowledge distillation, $N$ is the input one-hot dim, $p_i$ is the teacher probabilities, and $q_i$ is the student probabilities. The second term of the loss function is the mean squared error (*MSE*) of teacher's hidden variable $z_q$ and the student's hidden variable $z_p$. The third term of the loss function makes the output unquantized code, $z_e$, as close as possible to the quantized code, $e$. $sg$ is the stop gradient function, as the codebook ($e$) is updated by the Exponential Mean Sliding (EMA)[23]. $M$ is the codebook size (mol-gene length) and $\beta$ is the hyperparameter of the quantized loss factor.

$$Loss_{D-VAE_{encoder}} = \alpha * \left(-\frac{1}{N}\sum_{i=1}^{N} p_i^T \log q_i^T\right)$$
$$+ (1-\alpha) * MSE(z_q, z_p) + \beta * \frac{1}{M}\sum_{j=1}^{M}(z_e^j - sg[e_j])^2 \tag{3}$$

### Gene vector operator

In the mutation operation, the molecule is first converted to mol object using RDKIT[34], then randomly encoded into 3 different SMILES (the same molecule can have multiple different SMILES, but only one canonical SMILES), and then encoded into SELFIES, and finally encoded into mol-gene using D-VAE. The $\varphi * M$ positions of mol-gene were randomly selected to mutate to 0, and $\varphi$ is a preset coefficient of variation. The crossover operation first assigns weights based on the competition score of molecules in the population, and the weights of the molecules are based on the values after normalizing the scores to the range [0,1]. Then samples the two parent sets $A$ and $B$ based on the weights. Specifically, the population of molecules is divided into two sets, and then the molecules with weights greater than $\varepsilon$, a random number sampled from the standard normal distribution, are retained. Mol-genes in the parent sets are obtained in the same way as the mutation. For each mol-gene in the parent set $A$, $\varphi * M$ positions are randomly selected to be replaced with the value of the corresponding position in the parent set $B$.

### Selection algorithm

The initial value of the molecule's selection score (S-score) is 1. After a series of conditions are computed, the molecule is discarded if the S-score is less than the random value $\varepsilon$. QED is calculated based on the RDKIT built-in method and SA is calculated as reported[30]. Ancestral similarity is calculated from core and group sets. The core is obtained by extracting the scaffold of each elite molecule with MurckoScaffold and then calculating the MCS of the elite molecules with rdFMCS[34]. The set of groups is represented in the form of molecular fingerprints, and the set of motifs is obtained by vectorizing or calculating on all molecular fingerprints.

### Docking and Molecular Dynamic

The DGMM algorithm uses the Glide program with HTSV precision[32]. Autodock Vina is also supported[43]. In docking program, the molecules were input in SMILES format, and molecular conformations were generated using the ligprep program. Docking was performed using the protein grids based on the reported crystal structure of the proteins[44]. Molecular dynamics simulation (MD) was then conducted, followed by clustering to select the structure that best fits the positive drug[37]. DGMM also supports ensemble docking, which involves molecular docking based on multiple protein structures. To predict protein-ligand interaction during compound validation, induced fit docking (IFD)[38] combined with MD was used. The protein and protein-ligand complex structures were optimized using the same MD protocol, with 120 ns of arithmetic at 300 K in the NPT system.